\documentclass[runningheads]{llncs}
\usepackage{graphicx}
\usepackage{bm}
\usepackage{times}

\usepackage{soul}
\usepackage{url}
\usepackage[hidelinks]{hyperref}
\usepackage[utf8]{inputenc}
\usepackage[small]{caption}
\usepackage{graphicx}
\usepackage{amsmath}
\usepackage{booktabs}
\begin{document}
\title{Finding Beautiful and Happy Images for Mental Health and Well-being Applications\thanks{Supported by Guangdong Key Laboratory of Intelligent Information Processing, Shenzhen, China, Shenzhen Institute for Artificial Intelligence and Robotics for Society, Shenzhen, China, and Shenzhen Key Laboratory of Digital Creative Technology, Shenzhen, China.}}
%
%
\author{Ruitao Xie\inst{1} \and
Connor Qiu\inst{2} \and
Guoping Qiu\inst{3}}
\authorrunning{Xie et al.}
%
\institute{Shenzhen Institute of Advanced Technology, Chinese Academy of Sciences, Shenzhen, 518055, China, and University of Chinese Academy of Sciences, Beijing, 100049, China,
\email{rt.xie@siat.ac.cn}\\
\and
School of Public Health, Faculty of Medicine, Imperial College London, SW7 2AZ, UK,
\email{c.qiu@imperial.ac.uk}\\
\and
Shenzhen University, Shenzhen, China and Pengcheng National Lab, Shenzhen, China (on leave from the University of Nottingham, UK, \email{guoping.qiu@nottingham.ac.uk})}
\maketitle              
\begin{abstract}
This paper explores how artificial intelligence (AI) technology can contribute to achieve progress on \textbf{good health and well-being}, one of the United Nations’ 17 Sustainable Development Goals. It is estimated that one in ten of the global population lived with a mental disorder. Inspired by studies showing that engaging and viewing beautiful natural images can make people feel happier and less stressful, lead to higher emotional well-being, and can even have therapeutic values, we explore how AI can help to promote mental health by developing automatic algorithms for finding beautiful and happy images.  We first construct a large image database consisting of nearly 20K very high resolution colour photographs of natural scenes where each image is labelled with beautifulness and happiness scores by about 10 observers. Statistics of the database shows that there is a good correlation between the beautifulness and happiness scores which provides anecdotal evidence to corroborate  that engaging beautiful natural images can potentially benefit mental well-being. Building on this unique database, the very first of its kind, we have developed a deep learning based model for automatically predicting the beautifulness and happiness scores of natural images. Experimental results are presented to show that it is possible to develop AI algorithms to automatically assess an image's beautifulness and happiness values which can in turn be used to develop applications for promoting mental health and well-being.

\end{abstract}

\section{Introduction}
This paper explores how artificial intelligence (AI) technology can contribute to achieve progress on \textbf{Good Health and Well-being}, one of the United Nations’ 17 Sustainable Development Goals (SDGs). According to the latest available data\footnotemark[1] \footnotetext[1] {https://ourworldindata.org/mental-health}, it was estimated that 792 million people, i.e., more than one in ten of the global population (10.7\%) lived with a mental health disorder. The COVID-19 pandemic is exacerbating the mental health problem across the world. Developing AI solutions to improve mental health is therefore one of the most meaningful ways to achieve \textbf{AI for Good }.

Engaging and viewing beautiful natural images can make people feel happier and less stressful \cite{van2015autonomic}, lead to higher emotional well-being \cite{zhang2014engagement}, and can even have therapeutic values \cite{vincent2009therapeutic} \cite{reynolds2020virtual}. However, in these previous studies, the subjectively \textit{beautiful} images had to be chosen manually which means that it is very difficult to obtain large amount of such images to tailor the tastes of different users to benefit the mass. We therefore ask the question whether it is possible to develop AI solutions to automatically search for beautiful natural images that can make people feel happy and relaxing, thus enabling the development of applications for promoting mental health and well-being. The subject matters of this study is highly interdisciplinary and even the definitions of \textit{beautiful} and \textit{happy} images are very subjective concepts. These factors have made our attempt, the very first of its kind, to develop AI solutions to automatically search for beautiful and happy images with the expressed goal of promoting mental health, extremely challenging. Despite the difficulties, we have nevertheless developed a possible solution and make the following contributions.

Firstly, we contribute a large high quality Beautiful Natural Image Database (BNID) consisting of 20,996 very high resolution natural images. Each image in the BNID is labelled with a \textit{beautifulness} score and a \textit{happiness} score obtained from ratings by about 10 human observers. This is the first ever such database designed for developing automatic algorithms to search for beautiful and happy images for mental health applications. To ensure high image quality and diversity, the data was obtained by manually and algorithmically sieving through an initial set of 500,000 images down to a subset of 20,996, which were then rated by more than 200 human observers. Nearly 420,000 ratings were then obtained in this database, which have been made publicly available\footnotemark[2] \footnotetext[2]{https://drive.google.com/drive/folders/1qJ56Cvd5K4TR7NKBsW0kvu0iYu5LpUJx?usp=sharing} for research purposes.
We present statistics to show that in general there is a good positive correlation between the images beautifulness and happiness scores. For large majority of the cases, a beautiful image is also a happy image, and vice versus.  We also show that observers personal attributes and the ratings they gave exhibits some interesting patterns such as more optimistic persons will give higher beautifulness and happiness scores, the more pro-society the observers are, the higher scores they will give the images, and the better mood the observers are in, their scores are higher. Our BNID is the largest database providing anecdotal evidences which corroborate previous findings that beautiful natural images positively correlate with higher emotional well-being.

Secondly, we present a deep learning based system for automatically predicting images beautifulness scores and happiness scores. We employ content-based image retrieval to first find a similar image with manually annotated scores to provide a reference to improve prediction accuracy. Instead of predicting the beautifulness or happiness independently, we exploit the correlations between beautifulness and happiness by making use of an image's happiness score to assist the prediction of the beautifulness score, and vice versus, by  making use of an image's beautifulness score to assist the prediction of happiness score. We present experimental results to show that it is possible to automatically predict the beautifulness and happiness scores of natural images, which in turn can be used to automatically search for beautiful and happy images to be used for the purpose of developing applications aiming for improving mental well-being.

It is worth mentioning that the contribution of this paper is an initial attempt to explore how computer vision and pattern recognition algorithms can be used to automatically find beautiful and happy images for mental heath and well-being applications. The emphasis is on AI for Good rather than on developing the most advanced machine learning algorithms.

\section{Related work}
\label{Background}

The \textit{beautifulness} of an image is related to image aesthetics assessment (IAA) \cite{AestheticProcess2004}. Image aesthetic is very subjective and to complicate things further, a piece of work with a high aesthetic value can incur negative emotion. The \textit{happiness} of an image is related to affective image quality assessment \cite{AffectiveImage2010}. In this paper, we are only interested in image aesthetics that invokes positive emotions. The emotions evoked by a picture can be modelled in either positive or negative categorical emotion states \cite{achlioptas2021artemis}. 
Experiencing the aesthetics of beautiful pictures has been shown to invoke positive emotions and promote physical and physiological well-being \cite{zhang2014occasion,AestheticsLinkingWellbeing2019}. Positive emotion categories such as amusement, contentment, and excitement are finer grained versions of \textit{happiness} \cite{PositiveEmotionLinkingHappy2009}.

Early IAA algorithms mainly used hand-crafted features for rating image aesthetics. The advancement of deep learning has witnessed rapid increase in works based on deep neural networks.
Wang et al.~\cite{wang2016brain} leveraged popular neural networks for predicting image aesthetics.
Jin et al.~\cite{jin2019ilgnet} proposed a new neural network to extract the local and global features for image aesthetics assessment.

Similar to  IAA, early image emotion assessment (IEA) algorithms also used hand crafted features.
Multi-level region based  CNN has been developed for image emotion classification in \cite{RAO2019429}. For more comprehensive review and recent advances in the field IEA, readers are referred to recent surveys in \cite{zhao2021computational}.

\label{Aesthetics & Emotions} 
In fields outside computational sciences, researchers have long recognised the benefits of engaging with beautiful natural environments and studied how they can induce aesthetic and affective responses in humans \cite{Ulrich1983}. In fact, studies have shown that humans can engage with the so called \emph{technological nature} \cite{TechnologicalNature2009}, electronic display of natural environments, to gain emotional, psychological and physiological benefits. 

Research has shown that engaging and viewing beautiful natural images can make the viewers feel happier and less stressful \cite{van2015autonomic}, which in turn leads to higher emotional well-being~\cite{zhang2014engagement}, pro-social behaviours~\cite{zhang2014occasion}, higher levels of life satisfaction and a better spiritual outlook \cite{diessner2008engagement}. In addition to relieving stress, viewing beautiful natural images can also elicit improvements in the recovery process following acute mental stress \cite{zhang2014occasion,brown2013viewing}, reduce perception of pain~\cite{vincent2009therapeutic},  significantly reduce pro re nata medication incidents, help reducing mental health patients’ anxiety and agitation in healthcare settings \cite{nanda2011effect}, and can be a cost-effective treatment strategy for individuals with substance use disorder~\cite{reynolds2020virtual}. 

\section{A beautiful natural image database (BNID)}
\label{CAEdatabase}

\subsection{Collecting the images}
We started by collecting about 500,000 landscape images from the web. Considering aspect ratio can interfere with aesthetic rating, we only selected those images with an aspect ratio of about 3:2.
As images of low resolution will affect the reliability of rating, only those with resolutions nearly or exceeding 1800 $\times$ 1200 were kept. After first round of data cleaning, about 70,000 very high quality landscape images were left.
Content diversity and variety are important in order to have a high quality database. We calculated the color histograms of each image and compared the similarity between two images based on the distance of their color histograms. If the distance is below a preset threshold, we randomly throw away one of them. Through this round of data cleaning, 20,996 images were left for collecting beautifulness and happiness labels.

\subsection{Collecting beatutifulness and happiness  scores} 
In order to collect beautifulness and happiness scores, we built a website and recruited more than 200 volunteers.
We followed a common practice in the psychology literature \cite{zhang2014occasion} and adopted a 7-point scoring scale for both beautifulness and happiness scores. We used very simple and general language to instruct the participants to score the images. For collecting the beautifulness score, we presented the image to the raters along with the text: \emph{This image is beautiful, strongly agree = 7, strongly disagree = 1}. The raters are free to choose any one of the 7 scores. For collecting the happiness score, we presented the viewers with the image and the text: \emph{This image makes me feel happy, strongly agree = 7, strongly disagree = 1}. Again, the raters can choose any one of the 7 scores. 

Apart from these simple instructions, the raters have not received any training, they rated the images in their own environment and in their own time. To ensure reliability, the raters had to stay on one image for at least 3 seconds before moving to the next one. The raters participated in the exercise anonymously but we have asked them to fill in a questionnaire consisted of questions such as their gender, knowledge about photography, the lighting conditions of the viewing environment, and some personal attributes such as their mood when rating the images, their personality, general attitude towards society and their environments.     

\subsection{Analysis of beautifulness and happiness scores}
\begin{figure}
	\centering
	\includegraphics[scale=0.4]{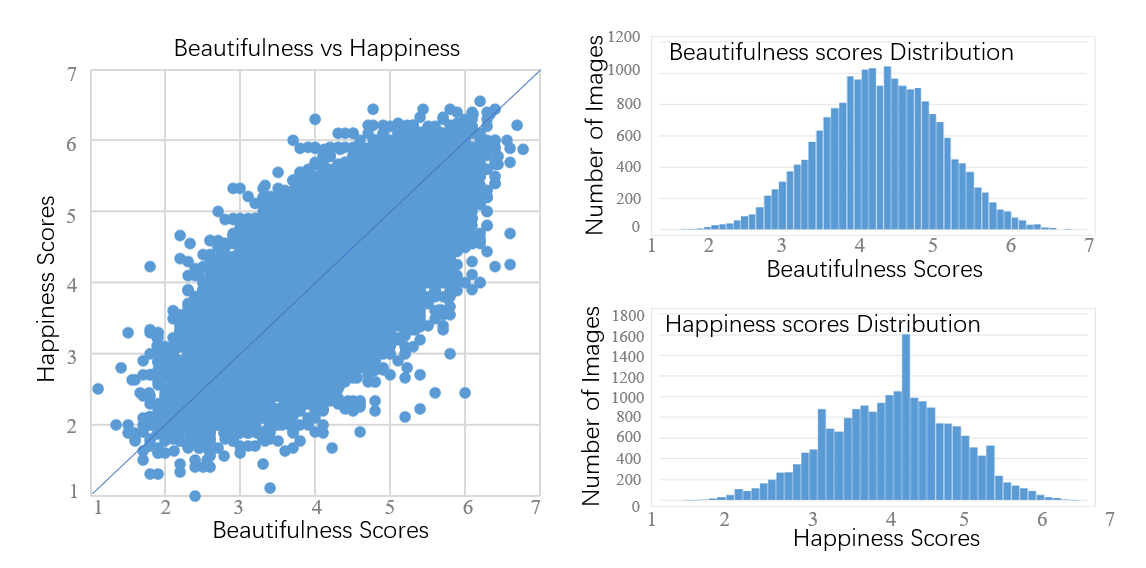}
	\caption{Left: Scatter plot between happiness and beautifulness scores. Right: Beautifulness and happiness score distributions}
	\label{scatter}
\end{figure}

It will be interesting to see how the images beautifulness scores relate to their happiness scores. Figure \ref{scatter} shows the relation between beautifulness and happiness scores and their distributions. It is seen that the distributions both have a bell shape. The beautifulness scores and happiness scores are highly correlated.

Table \ref{distribution_between_aa} shows the beautifulness-happiness difference distribution. It is seen that for the majority of the images, their beautifulness and happiness scores only differ between 0 and 0.5.
These statistics show that in majority of the cases, there is a strong positive correlation between an image's beauty and the positive emotion (happiness) it invokes, and that a beautiful image is also a happy image, and vice versus.  It is important to note that all image in the BNID are natural landscape images, these relations may not hold for arbitrary contents. Indeed, all conclusions in this paper are based on such image types.
\begin{table}
	\centering
	\caption{Beautifulness-Happiness Difference Distribution}
	{\begin{tabular}[l]{@{}lccccc} 
			\toprule
			Diff. interval&$<0.25$&$<0.5$&$<0.75$&$<1$\\
			\midrule
			No of Images &30.62\% &54.05\% & 75.71\% & 86.95\%		
	\end{tabular}}
	\label{distribution_between_aa}
\end{table}

\begin{figure}
	\centering
	\includegraphics[scale=0.32]{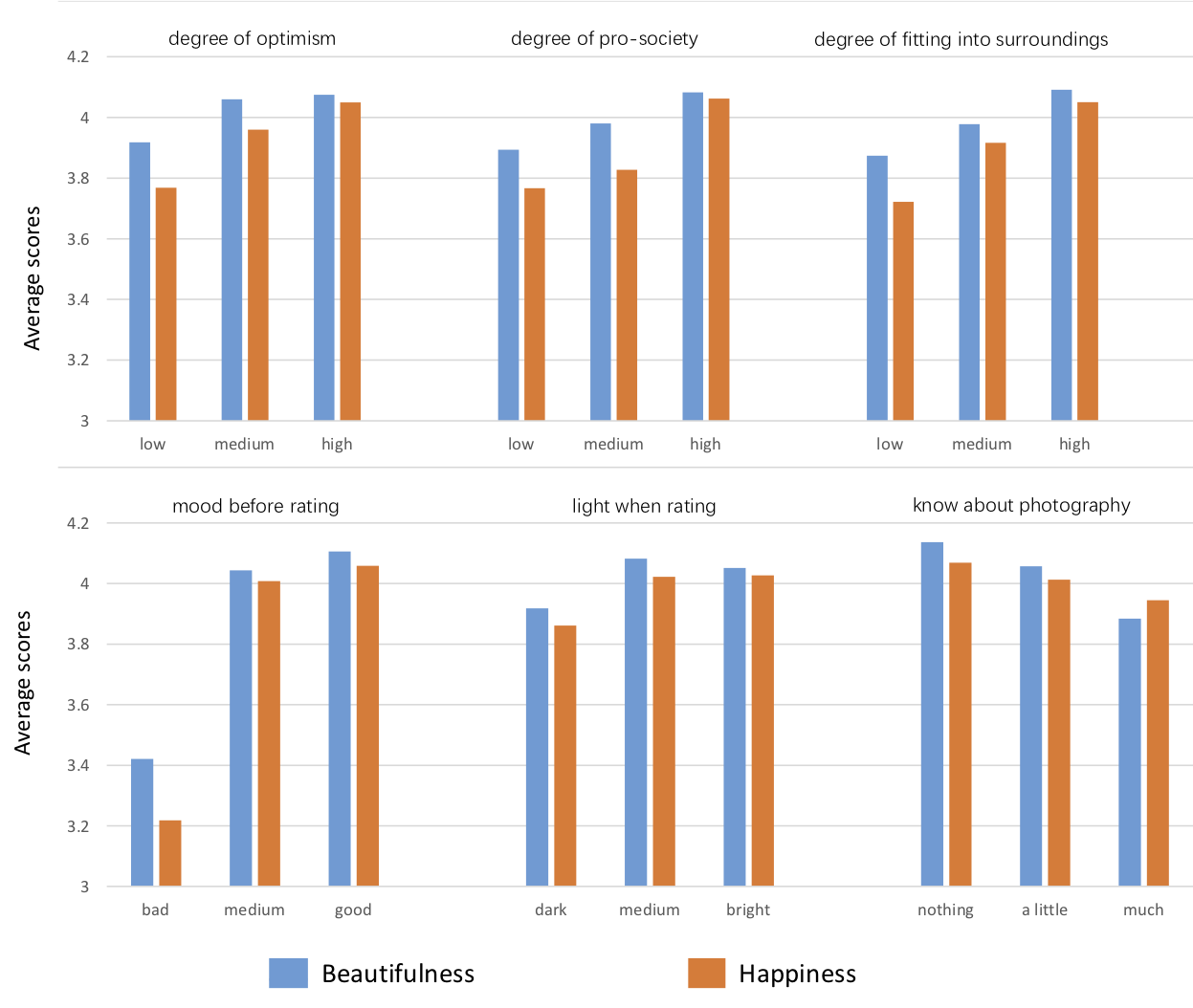}
	\caption{Observers personal attributes and other factors versus raters' average beautifulness and happiness scores.}
	\label{person}
\end{figure}

How raters' personal attributes and the viewing environment affect their ratings is of interest. Figure~\ref{person} show how raters' degrees of optimism, pro-society, fitting with the surrounding environments, their mood before started the rating exercise, their knowledge of photography, and the lighting condition of the viewing environment affect their scores. It is very interesting to observe that more optimistic persons will give higher beautifulness and happiness scores, the more pro-society the raters are, the higher scores they will give the images, the better the raters can fit in with their surroundings, the higher the scores they will rate the images, and the better mood the raters are in, their scores are higher. In terms of the viewing environment, a lighter environment tend to make raters give higher scores. Not surprisingly, the more knowledge the raters had about photography, the lower scores they gave. This is because they may judge the photos from the rules of photography rather than the beauty of the images.  

These results supports findings from previous studies that engaging with beautiful natural images can gain emotional and well-being benefits \cite{van2015autonomic}, \cite{zhang2014engagement}, \cite{zhang2014occasion},  \cite{brown2013viewing}, \cite{vincent2009therapeutic}, \cite{nanda2011effect} and \cite{reynolds2020virtual}. In order to enable the public to benefit from such beautiful natural images, we need to be able to automatically find large amount of such images to engage the public to help improve the mental well-being.    

\section{Beautifulness  and happiness assessment }
\begin{figure}
	\centering
	\includegraphics[scale=0.2]{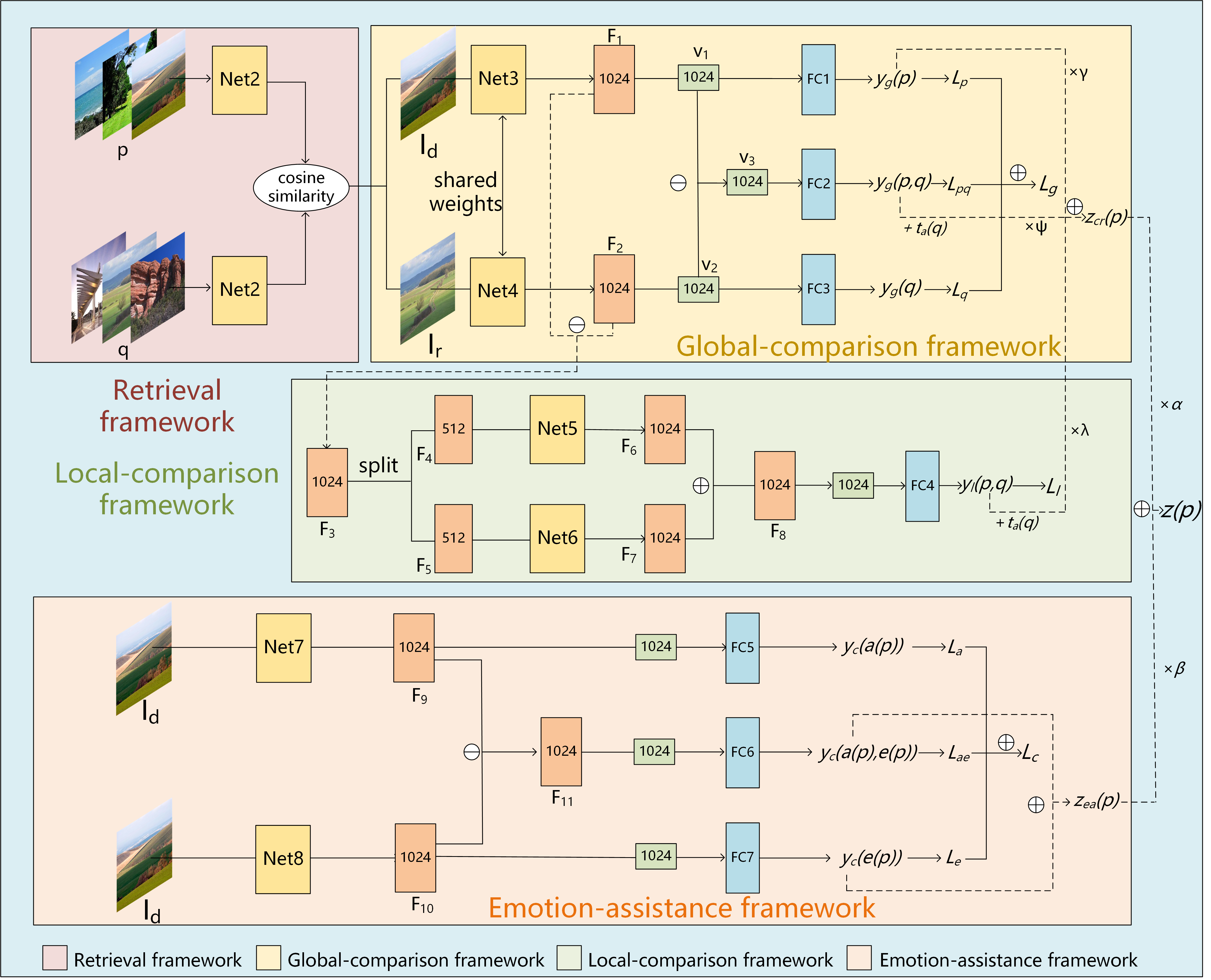}
	\caption{The overall framework of our proposed algorithm for image beautifulness assessment. In all modules, the yellow boxes refer to the networks used, the brown and green boxes represent the features, in which the numbers indicate the channels of features, and the blue boxes represent the classifiers composed of fully connected layers. $\bm{p}$ is the input image and $\bm{q}$ is the paired reference image for comparison.}
	\label{fig:overall-framework}
\end{figure}

We begin by describing our method for image beautifulness assessment which can be easily extended to image happiness assessment. 
The overall image beautifulness assessment framework is shown in Figure~\ref{fig:overall-framework} which consists of 4 major components.
The retrieval module uses content-based image retrieval (CBIR) \cite{Dubey2020ADS} to retrieve a reference image that is similar to the input. The global comparison module consists of a Siamese network that takes the input image and its reference as input to predict the input image's beauty score. The local comparison module explicitly makes use of the local features to assist beauty prediction.
The emotion-assistance module predicts the input image's happiness value and its relation with the beautifulness of the input image to assist the prediction of its beauty value.

\subsection{Image beautifulness assessment}
Instead of evaluating an image's aesthetics (beautifulness) in isolation, we introduce a content-referenced strategy to give the prediction model contextual information to enhance the stability and reliability of image beautifulness assessment. We exploit a Siamese network for feature extraction to ensure that the same set of features are extracted from the input and its reference.
The Siamese network is based on Densenet121~\cite{huang2017densely} pre-trained on ImageNet. This network architecture integrates features of different layers and can better integrate semantic and structural information to construct powerful representative features to facilitate beauty score prediction. 

Local information is also important for image aesthetics assessment~\cite{lu2014rapid}. Various local information will have different effects on image beauty, the aesthetics of the central area will largely determine the image's beauty. To leverage local information to achieve better beautifulness assessment performances, we introduce a module to estimate the relative aesthetic score between the input image and its reference image.

As discussed previously and demonstrated in Section \ref{CAEdatabase}, the beautifulness and happiness scores of an image are highly correlated.
In order to make full use of the happiness information of the images,
we propose a collaborative beautifulness and happiness assessment solution. As shown in the emotion assistance module in Figure~\ref{fig:overall-framework}, we use two parallel networks to regress the beautifulness score and the happiness score respectively. The feature map $F_{9}$  represents aesthetic related image features while $F_{10}$ contains happiness related features. The feature map $F_{11}$ which is obtained by subtracting $F_{10}$ from $F_{9}$ mixes the beautifulness and happiness features and is used to estimate the difference between the beauty and happiness scores of the image.
A one-layer classifier is used to predict the beautifulness and happiness scores respectively whilst a two-layer classifier is used to predict the difference between the scores. Similarly, DenseNet121 pre-trained on ImageNet is used to implement the feature extraction here.

Beautifulness and happiness scores are content dependent. It is more meaningful and reliable to compare two images with similar contents. We use content based image retrieval (CBIR) \cite{Dubey2020ADS} to find a similar image from the training set as a reference image to help build better and more reliable image beautifulness assessors. The retrieval process leveraged the features extraction module of Densenet161~\cite{huang2017densely} pre-trained on the Places365 dataset~\cite{zhou2017places} used for scene classification. The feature maps of input image $p$ and reference image $q$ extracted from the pre-trained Densenet161 are used to derive two feature vectors $\bm{v_{p}}$ and $\bm{v_{q}}$ using the average pooling operator. The distance between $\bm{v_{p}}$ and $\bm{v_{q}}$ are then used to find the most similar image from the reference image set as the input image's reference. 

\subsection{Loss functions}
Training of the overall framework is divided into 3 stages corresponding to the training of the global, local and emotion-assistance modules. As shown in equations (\ref{con:Loss2}) to (\ref{con:Loss4}), we use mean square errors to design the loss functions.
\begin{equation}
	\begin{aligned}
		\mathcal{L}_{g}=&(y_{g}(p)-t_{a}(p))^2+ (y_{g}(q)-t_{a}(q))^2 \\&+ (y_{g}(p,q)-(t_{a}(p)-t_{a}(q)))^2
	\end{aligned}
	\label{con:Loss2}
\end{equation}	
\begin{equation}
	\mathcal{L}_{l}=(y_{l}(p,q)-(t_{a}(p)-t_{a}(q))^2
	\label{con:Loss3}
\end{equation}
\begin{equation}
	\begin{aligned}
		\mathcal{L}_{c}=&(y_{c}[a(p)]-t_{a}(p)^2+(y_{c}[e(p)]-t_{e}(p)^2\\&+(y_{c}[a(p),e(p)]-(t_{a}(p)-t_{e}(p))^2
	\end{aligned}
	\label{con:Loss4}
\end{equation}

where $t_{a}(p)$ and $t_{e}(p)$ respectively represent the ground truth of the beautifulness and happiness scores of the input image $p$, and $t_{a}(q)$ is the beautifulness ground truth of the reference image $q$, $y_{g}(p)$ and $y_{g}(q)$ are respectively the predicted beauty scores of the input image $p$ and reference image $q$ by the global comparison module, $y_{g}(p,q)$ and $y_{l}(p,q)$ are respectively the predicted relative beautifulness values between images $p$ and $q$ by the global comparison and the local comparison modules, $y_{c}[a(p)]$ and $y_{c}[e(p)]$ are respectively the predicted beautifulness score and happiness score of image $p$, and $y_{c}[a(p),e(p)]$ is the predicted difference between beautifulness and happiness of image $p$.

\subsection{Final score}

Empirically, we found that by combining predictions from the content-referenced module and the emotion assisted module gives the best results. The final beautifulness score of image $p$, $z(p)$ is estimated as follows
\begin{equation}
	z(p) = z_{cr}(p)+ z_{ea}(p)
	\label{con:final_result}
\end{equation}
where $z_{cr}(p)$ and $z_{ea}(p)$  are respectively the beautifulness scores of image $p$ estimated by the content-referenced module and the emotion assisted module, $z_{cr}(p)$ and $z_{ea}(p)$ are calculated as follows   
\begin{equation}
	\begin{aligned}	
		{z_{cr}(p)= y_{g}(p)+y_{g}(p,q)+y_{l}(p,q) + 2t_{a}(q)}
	\end{aligned}
	\label{con:aes_result}
\end{equation}
\begin{equation}
	z_{ea}(p) = y_{c}[e(p)]+y_{c}[a(p),e(p)]
	\label{con:emo_result}
\end{equation}

\subsection{Extension to image happiness prediction}

We can easily extend the framework in Figure \ref{fig:overall-framework} to predict images' happiness score. All we have to do is to substitute beautifulness by happiness and happiness by beautifulness in Figure\ref{fig:overall-framework} and in equations (\ref{con:Loss2}) to (\ref{con:emo_result}). As we shall show in the experimental section, this is indeed the case. 

\section{Experiments}
In order to ensure data quality for effectively verifying our automatic assessment algorithms, we removed noisy ratings from the data by first removing outlier scores of each image and then deleting images that have too few valid ratings after outlier removal. After this round of data cleaning, 14644 images were used in our experiments.
We randomly sample 2218 images with beautifulness scores between 3.8 and 4.2 as the reference images, in which 1663 are used during training and 555 are used in testing. The remaining 12426 are used as input images in which 9320 randomly sample are used for training and the other 3106 are used for testing.
To reduce over fitting, we perform data augmentation by resize the images by some random factors and flipping horizontally with a probability of 0.5.

\textbf{Evaluation metrics}.
We divide images in the database into two classes. Those with scores exceed or equal to $4$ are classified as good and the others are defined as bad.
We calculate the accuracy ($ACC$) based on binary classification (good or bad) and use it as one of the evaluation criteria. Furthermore, we take the mean square error ($MSE$) between the ground truth and the predicted results as another evaluation metric. Spearman's rank correlation coefficient ($SRCC$) and linear correlation coefficient ($LCC$) are also computed between the predicted and ground truth scores.

\subsection{Image beautifulness  assessment results}
As a comparison baseline, we fine tune a Densenet121 end-to-end, input an image and output its corresponding beautiful score as is done in the vast majority of existing deep learning based approaches. 

\begin{figure}
	\centering
	\includegraphics[scale=0.26]{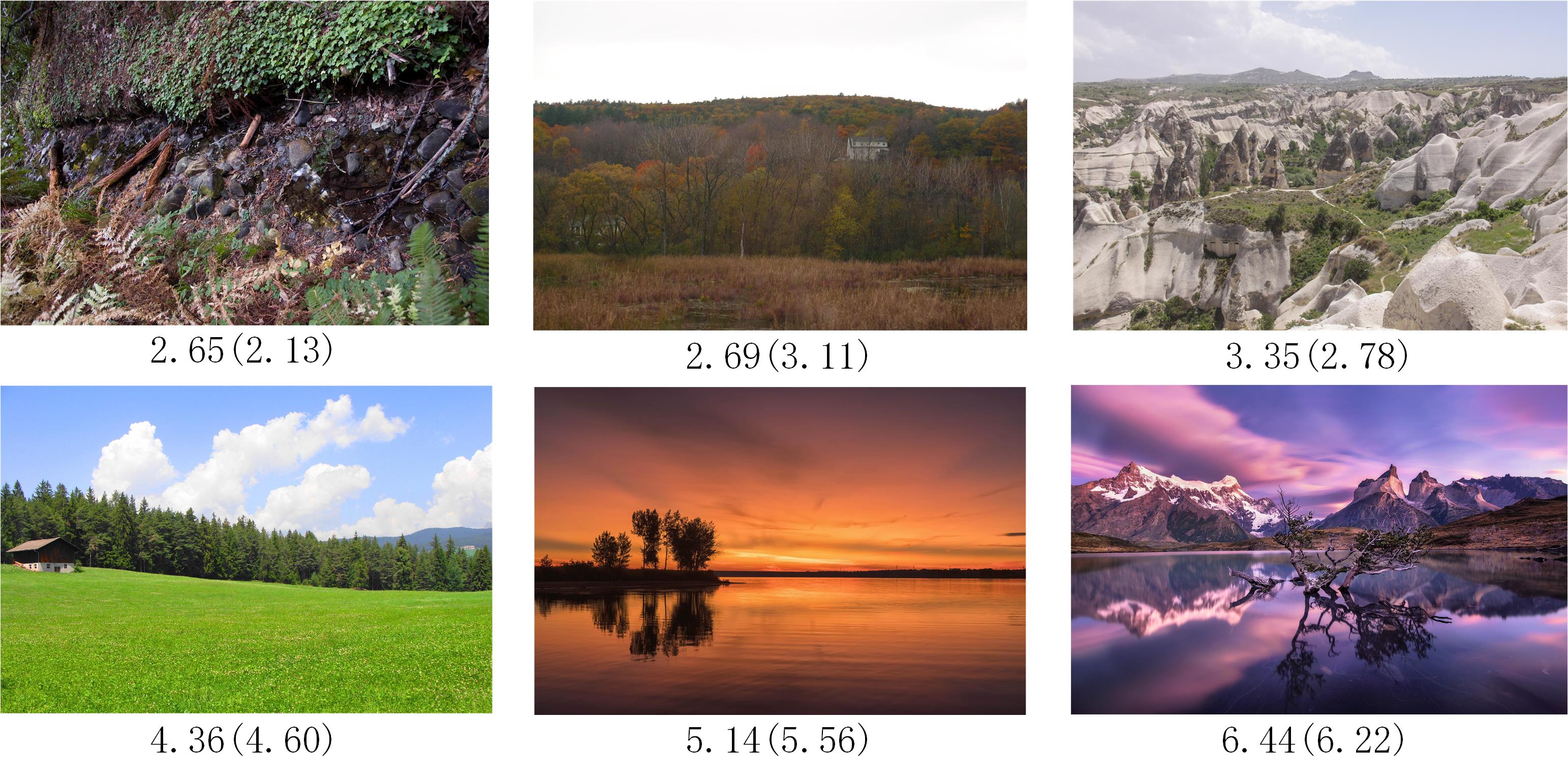}
	\caption{Predicted beautifulness scores. The number below are predicted scores and ground truth scores (inside bracket)}
	\label{fig:fianl-model-cases}
\end{figure}

\begin{table}
	\centering
	\caption{Results of beautifulness prediction}
	
	{\begin{tabular}[l]{@{}lcccc}
			
			\toprule
			
			Metrics&$ACC$&$MSE$&$SRCC$&$LCC$\\
			\midrule
			Baseline &73.79\% &0.7602 & 0.6331 & 0.6361\\		
			\multicolumn{5}{c}{Content reference without emotion assistance} \\
			wo/ local & 75.34\% & 0.7165 & 0.6525 & 0.6540  \\		
			w/ local & 75.82\% & 0.7024 & 0.6597 & 0.6631 \\
			wo/ retrieval & 74.66\% & 0.7241 & 0.6449 & 0.6486  \\
			\multicolumn{5}{c}{Content reference plus emotion assistance}\\
			w/ emotion (single)& 76.14\% & 0.6886 & 0.6663 & 0.6705  \\		
			w/ emotion (Full)  & \textbf{77.33\%} & \textbf{0.6547} & \textbf{0.6871} & \textbf{0.6903}  \\
			Improvement  &4.80\% & 13.88\% & 8.53\% & 8.52\%  \\
			\bottomrule
			
	\end{tabular}}
	
	\label{twin_network_design}
	
\end{table}
\textbf{Overall results}. The second last line in Table~\ref{twin_network_design}, w/emotion (Full), shows our full system's beautifulness score prediction result. 
We can see that with a complete content reference module and a complete emotion assisted module, accuracy ($ACC$) performance has reached to 77.33\%, achieving a 4.80\% improvement over the baseline. For the mean square error ($MSE$) metric, our approach has achieved a 13.88\% improvement over the base line. In the cases of Spearman's rank correlation coefficient ($SRCC$) and linear correlation coefficient ($LCC$) performances, the method has improved 8.53\% and 8.52\% respectively over baseline.
Figure~\ref{fig:fianl-model-cases} shows some example images and the predicted beautifulness scores by our method. As can be seen, for those images that are obviously beautiful, the predicted scores are high, and those obvious unattractive looking, the predicted scores are low. These examples clearly demonstrate the excellent ability of our method in assessing images' beauty values.

\textbf{Content referencing and CBIR}. In this paper, we introduce content reference and content-based image retrieval (CBIR) into image beautifulness assessment as shown in our overall framework in Figure \ref{fig:overall-framework}. Here we study the roles they play. We first investigate the roles of global and local features. As shown in Table \ref*{twin_network_design}, explicitly making use of local features can indeed help improve the system's ability in better predicting the beautifulness scores. It is seen that including local features (w/local)  is better than using global feature only (wo/local). To investigate the soundness of content reference and the importance of retrieving similar images based on CBIR, we perform experiments by randomly choosing reference images from the reference set. As seen in Table \ref*{twin_network_design}, even use randomly chosen reference image, the results are better than the baseline, demonstrating the usefulness of using content reference. Compared to results based on CBIR (indicated by "w/local" in the Table \ref*{twin_network_design}), using randomly chosen reference image performs worse which is entirely expected.

\textbf{Happiness helps beautifulness prediction}. Table \ref*{twin_network_design} also lists results when the emotion assisted module is included in the full system. The result labelled as "w/emotion (single) means when the happiness score of the input image is estimated by a single network (the Net8-$F_{10}$-$FC7$ path in Figure \ref{fig:overall-framework}) and replacing $Z_{ea}(p)$ with $y_c(e(p))$. 
It is seen that happiness can indeed assist the prediction of beautifulness.
The full result is when the complete emotion assistance module, where we not only learn the happiness score of the input image, but also its relation with the beautifulness is also explicitly modelled. This result shows that modelling happiness and beautifulness together through extracting happiness and beautifulness features and then mixing them together for learning the relation between the two can help improve performances. This further demonstrate the intrinsic relation between the two.

\begin{figure}[h]
	\centering
	\includegraphics[scale=0.26]{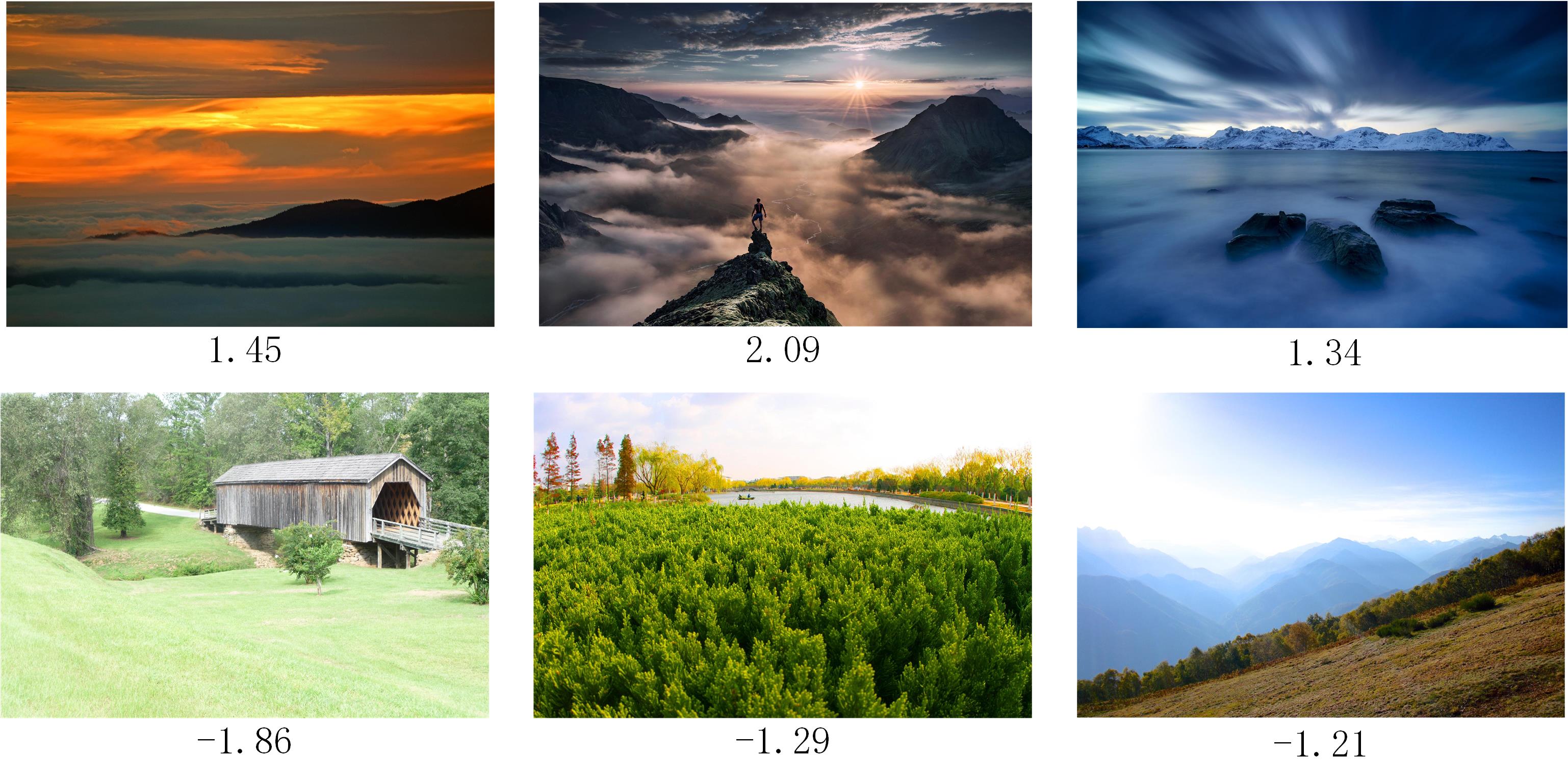}
	\caption{Example image and their predicted beautifulness-happiness difference. The numbers below each image represent predicted $beautifulness$ - $happiness$. A positive means the image's beauty score is higher than happiness score whilst a negative means means the image's happiness is higher than beautifulness.}
	\label{fig:EMAD-relation-diff-large}
\end{figure}

As mentioned previously and as can be expected that not all images' beauty scores are positively related with their happiness scores. There are cases where the gap between the scores can be large. It is for this reason, we introduce a network branch that models the difference between the beautifulness and happiness of the image in our emotion-assistance module (see the middle blocks in the emotion assistance module ).
Figure ~\ref{fig:EMAD-relation-diff-large} shows some example images and the predicted beautifulness-happiness difference of the images.
We can see that large difference values are predicted for those images which are good-looking but with dark tone (a dark tone is more likely to make the viewers feeling sad), or those images look bright (generally a bright tone can cheer people up) but with normal looking and not so beautiful contents. 
These examples also demonstrate the ability of our system in modelling the gap between an image's beautifulness and happiness scores.

\subsection{Image happiness assessment results}

\begin{figure}
	\centering
	\includegraphics[scale=0.25]{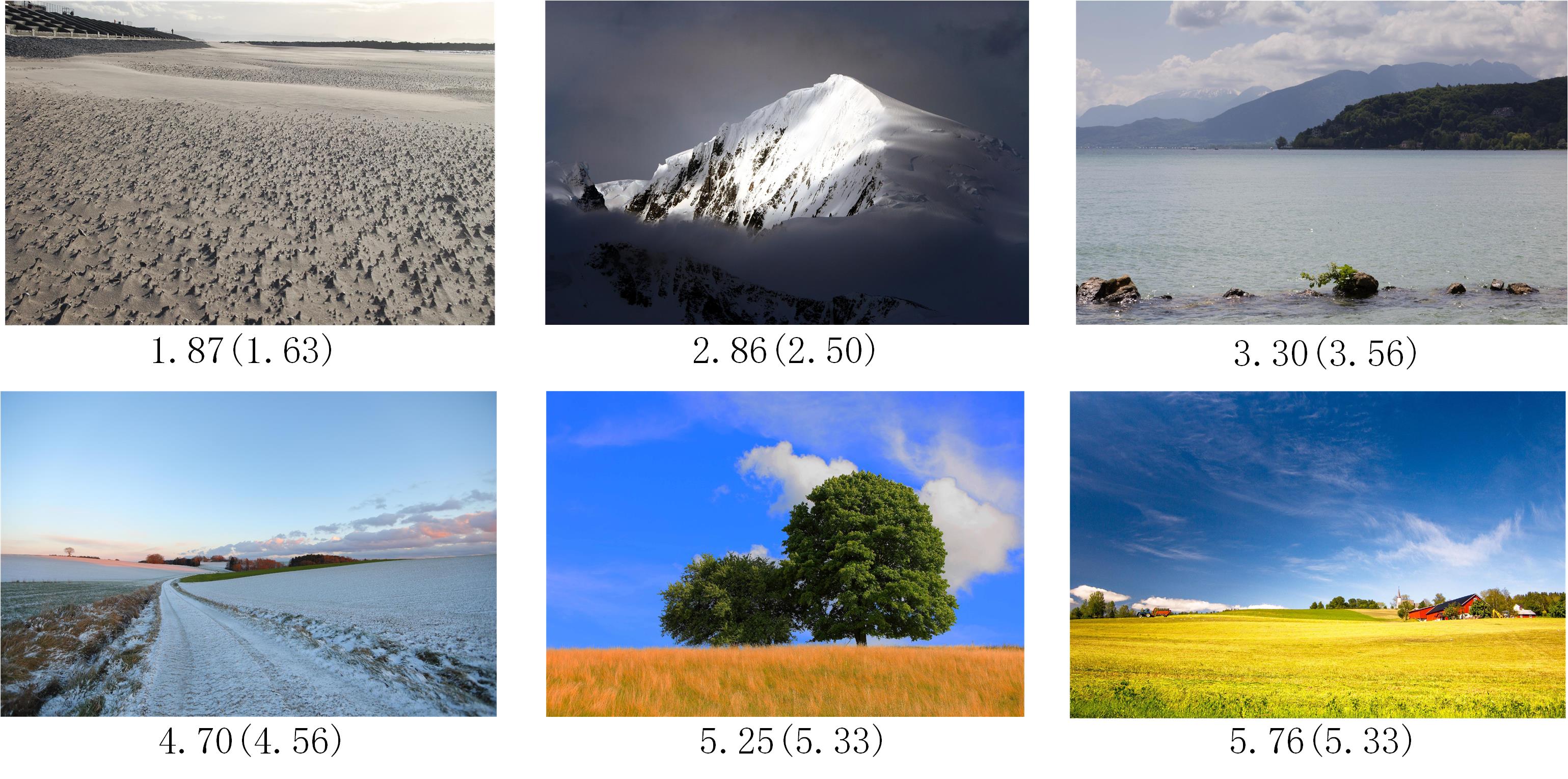}
	\caption{Predicted happiness scores. The number below are predicted scores and ground truth scores (inside bracket)}
	\label{fig:fianl-emo-model-cases}
\end{figure}

By swapping places between beautifulness and happiness, we can easily turn the scheme in Figure \ref{fig:overall-framework} to predict image happiness score. For the happiness assessment experiments, we also take the similar data split strategy and as a result, 9329 and 1654 images are randomly selected as the input and reference images respectively for training. The numbers of input and reference images in the test set are 3110 and 551 respectively. The emotion assessment algorithm implementation is the same as beautifulness prediction shown above (only swapping positions between beautifulness and happiness). The results of happiness assessment are presented in Table~\ref{emo_assessment_results}. As we can see, the content reference and aesthetics assistance methods also bring improvement in image happiness assessment. Figure~\ref{fig:fianl-emo-model-cases} shows some example images with their predicted happiness scores which further demonstrate the reliability of our proposed algorithms for happiness prediction.
\begin{table}
	\centering
	\caption{Happiness prediction results}
	
	{\begin{tabular}[l]{@{}lcccc}
			
			\toprule
			
			Metrics&$ACC$&$MSE$&$SRCC$&$LCC$\\
			
			\midrule
			
			Baseline &77.62\% &0.6643 & 0.7026 & 0.6963\\
			\multicolumn{5}{c}{Content reference without beautifulness assistance} \\
			wo/ local & 79.26\% & 0.6494 & 0.7119 & 0.7108  \\		
			w/ local & 79.45\% & 0.6421 & 0.7159 & 0.7141 \\
			wo/ retrieval & 78.78\% & 0.6535 & 0.7059 & 0.7085  \\
			\multicolumn{5}{c}{Content reference plus beautifulness assistance}\\
			w/ beauty (single)& 79.81\% & 0.6355 & 0.7265 & 0.7264  \\
			w/ beauty (Full)  &\textbf{80.42}\% & \textbf{0.5895} & \textbf{0.7450} & \textbf{0.7455}  \\
			Improvement  &3.60\% & 11.26\% & 6.03\% & 7.07\%  \\
			\bottomrule
			
	\end{tabular}}
	
	\label{emo_assessment_results}
	
\end{table}

\section{Concluding remarks}
Aiming at exploiting artificial intelligence to help improve mental health, this paper has attempted for the first time to study image beautifulness and happiness under a unified computational framework. We have constructed the first ever very high quality large natural image database in which each image is labelled with a beautifulness score and a happiness score. We show that natural landscape images beautifulness scores are highly correlated with their happiness scores, and that these scores supports previous findings that beautiful natural images can benefit emotional well-being. Building on this database, we have developed a deep learning system for automatically predicting the beautifulness and happiness scores of natural images. Such system can be used to search for beautiful and happy images for the purpose of developing applications that promote mental health and well-being.

\end{document}